\documentclass[10pt,twocolumn]{article} 
\usepackage{simpleConference}
\usepackage{times}
\usepackage{graphicx}
\usepackage{amssymb}
\usepackage{url,hyperref}

\usepackage{algorithm}
\usepackage{algorithmic}
\usepackage{booktabs}       
\usepackage{amsfonts}       
\usepackage{nicefrac}       
\usepackage{microtype}      
\usepackage{xcolor}         
\usepackage[utf8]{inputenc} 
\usepackage{comment}
\usepackage{textcomp}
\usepackage{xspace}
\usepackage{mathtools}
\usepackage{bbm}
\usepackage{makecell, verbatim, sidecap}
\usepackage{subfig}
\usepackage{multirow}
\usepackage{gensymb}
\usepackage{amsmath,amssymb} 

\def\ie{{\it i.e.}\xspace}

\def\ReLU{{\rm ReLU}}

\begin{document}

\title{Two-Stream Networks for Object Segmentation in Videos}

\author{
    Hannan Lu\textsuperscript{\rm 1},
    Zhi Tian\textsuperscript{\rm 2},
    Lirong Yang\textsuperscript{\rm 2},
    Haibing Ren\textsuperscript{\rm 2},
    Wangmeng Zuo\textsuperscript{\rm 1} \\
    \{hannanlu, wmzuo\}@hit.edu.cn,
    \{tianzhi02, yanglirong, renhaibing\}@meituan.com
}


\maketitle
\thispagestyle{empty}

\begin{abstract}
Existing matching-based approaches perform video object segmentation (VOS) via retrieving support features from a pixel-level memory, while some pixels may suffer from lack of correspondence in the memory (i.e., unseen), which inevitably limits their segmentation performance. 
In this paper, we present a \textbf{T}wo-\textbf{S}tream \textbf{N}etwork (TSN).
Our TSN includes 
(\textit{i}) a pixel stream with a conventional pixel-level memory, to segment the \textit{seen} pixels based on their pixel-level memory retrieval.
(\textit{ii}) an instance stream for the \textit{unseen} pixels, where a holistic understanding of the instance is obtained with dynamic segmentation heads conditioned on the features of the target instance.
(\textit{iii}) a pixel division module generating a routing map, with which output embeddings of the two streams are fused together. 
The compact instance stream effectively improves the segmentation accuracy of the \textit{unseen} pixels, while fusing two streams with the adaptive routing map leads to an overall performance boost.
Through extensive experiments, we demonstrate the effectiveness of our proposed TSN, and we also report state-of-the-art performance of 86.1\% on YouTube-VOS 2018 and 87.5\% on the DAVIS-2017 validation split.
\end{abstract}

\section{Introduction}

Video Object Segmentation (VOS) is a fundamental task in video analysis, and has been widely applied in computer vision applications, such as video content editing and automatic driving.
In recent years, semi-supervised VOS has been developed extensively and achieved great progress.
Its goal is to separate specific object(s) from the background for each frame in a video sequence, given the mask of the target object(s) in the initial frame.

\begin{figure}[t]
  \centering
  \includegraphics[width=0.46\textwidth]{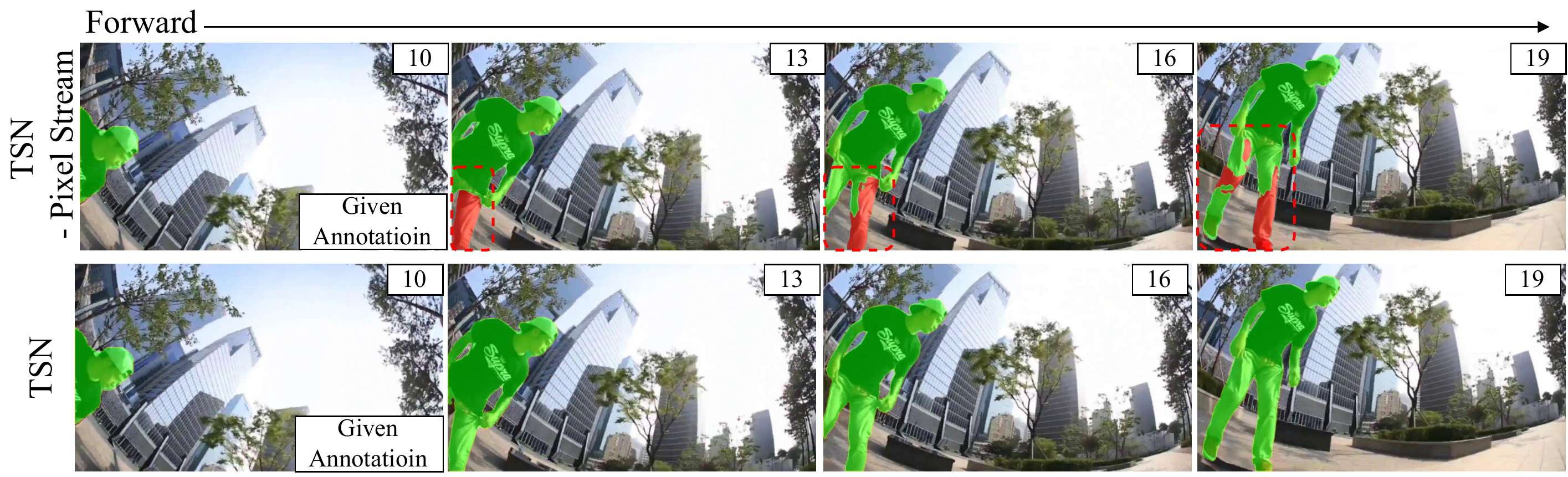} 
  \caption{
    Visualization of segmenting with a pixel stream alone. 
    %
    %
    In the forward pass, the pixel stream (top row) fails to segment the leg area correctly since it is unable to find a correspondence in the reference frame.
    Our two-stream network (bottom row) can address this issue effectively.
  }
  \label{figure_motivation}
\end{figure}

Top-performing works for semi-supervised VOS primarily exploit the template matching technique, in which labels are propagated from a reference set (\ie, the first frame with a given annotation and the historical segmented frames) to the query frame (\ie, the current frame) through matching.
Inspired from the memory network~\cite{MemoryNetwork}, STM~\cite{STM} constructs a pixel-level memory with multiple historical frames, where pixel-level features are extracted and stored in the memory.
Afterwards, the query pixels are segmented based on their retrieved reference features in the memory with their pixel-wise matching affinity.
Based on STM, most follow-up works\cite{KMN,LCM,HMMN,RMNet,STCN,SST,AOT} are devoted to investigating solutions to obtain more accurate retrieval from pixel-level memory.
However, these methods still have trouble in handling pixels not appeared in the reference frames (\ie, unseen), which lack a correspondence in the pixel-level memory.
As shown in Figure~\ref{figure_motivation}, such approaches relying on pixel-level memory fail to segment the emerging leg area due to its absence in the reference.

In this paper, we propose a \textbf{T}wo-\textbf{S}tream \textbf{N}etwork (TSN) to address this issue.
TSN mainly includes a pixel stream and an instance stream, which are fused with a pixel division module.
The pixel stream is responsible for the \textit{seen} pixels as in previous methods, and these pixels can often be accurately segmented because it is able to find similar correspondences in the pixel-level memory.
In contrast, the instance stream is mainly in charge of segmenting the \textit{unseen} pixels, which lack a correspondence in the pixel-level memory.
For segmentation, the instance stream generates segmentation heads composed of dynamical kernels, which are conditioned on the feature of each instance of the target object.
Integrating the instance-level information endows the dynamic kernels in the instance stream with a holistic understanding of the instance and reasoning ability, which makes it more suitable for segmenting the \textit{unseen} pixels.
To obtain the final segmentation results, the features of the two streams are weighted-aggregated with a routing map, which is generated with a proposed pixel division module and decide how much the a pixel should rely on the two streams, respectively.
We evaluate our TSN on two popular VOS benchmarks, \ie, DAVIS and YouTube, and report the new state-of-the-art results.
We summarize the main contributions as follows,
\begin{itemize}
\item We propose a two-stream network for semi-supervised VOS, which mainly includes a pixel stream, an instance stream and a pixel division module to fuse them together.

\item The proposed instance stream improves the segmentation accuracy of the unseen pixels, attributing to its dynamic convolution bank generated conditioned on the features of the target instance.
While the pixel division module adaptively merges the outputs of the two streams through the routing map, which effectively improves the overall performance.

\item TSN achieves the state-of-the-art results on two popular VOS benchmarks, \ie, DAVIS and YouTube-VOS.
\end{itemize}

\section{Related Work}

\noindent{\textbf{Semi-supervised video object segmentation}.}
Early semi-supervised video object segmentation (VOS) methods~\cite{OSVOS, OSVOS-S} apply segmentation-by-detection scheme, where general segmentation models are fine-tuned on the first annotated frame to focus on specific targets.
To tackle with the evolution of object appearance over time, OnAVOS~\cite{OnAVOS} online updates the object-specific model along the video.
For efficient inference, subsequent solutions~\cite{MaskRNN, MaskTrack, PReMVOS} resort to motion cues for temporal propagation, some of which~\cite{PReMVOS, LucidDreaming} propagates object mask from the first frame to the succeeding frames leveraging optical flow.

Recently, more researchers tend to formulate the semi-supervised VOS task as a template matching problem, where labels are propagated from a reference set to the query frame through pixel-wise matching and retrieval.
Prior works~\cite{shin2017pixel, videomatch} employs siamese structure network to extract features from both template frames and the query frame.
VideoMatch~\cite{videomatch} updates the template feature set with segmented frames.
Under the assumption of temporal continuity, FEELVOS~\cite{FEELVOS} narrows the matching area with the latest frame to a local range.
CFBI~\cite{CFBI} further promotes FEELVOS~\cite{FEELVOS} by establishing template sets for foreground and background respectively.
Utilizing memory network~\cite{MemoryNetwork, NeuralTuringMachines, metaMemory}, STM~\cite{STM} constructs spatio-temporal memory with multiple segmented frames.
The success of STM greatly encourages the development of matching based solutions~\cite{EGMN, GCM, AFB-URR, swiftnet, KMN, RMNet, LCM, HMMN}.
EGMN~\cite{EGMN} develops a graph structure memory network, where memory reading and writing are performed on each node sequentially. 
STCN~\cite{STCN} improves matching accuracy applying the negative squared Euclidean distance as similarity metric.
Later works~\cite{SST, AOT, Joint} introduce transformer~\cite{transformer} to VOS, where matching and retrieval are carried out multiple times with multiple attention layers.
Joint~\cite{Joint} complements pixel-level retrieval with an online updated discriminative branch additionally.

The aforementioned solutions classify pixels to foreground/background based on their pixel-level retrieval.
However, pixels not appeared in the past frames emerge along the video, which may be unable to find correspondence in the reference.
We propose to disentangle the query pixels with a routing map calculated based on the residual of the feature of query and their corresponding retrieval.
The value of residual indicates whether pixels can find similar correspondence in the reference.

\noindent{\textbf{Instance-level segmentation}.}
Through global pooling the feature of all foreground pixels, CFBI and RPCM~\cite{CFBI, RPCM} generate object-level embedding for channel re-weighting on query feature.
For instance segmentation, CondInst~\cite{CondInst} generates dynamic segmentation heads for each instance.
In this work, we construct an instance-level memory with segmentation heads dynamically generated integrating each instance of the target object.

\section{Our Approach}

\begin{figure*}[t]
  \centering
  \includegraphics[width=0.99\textwidth]{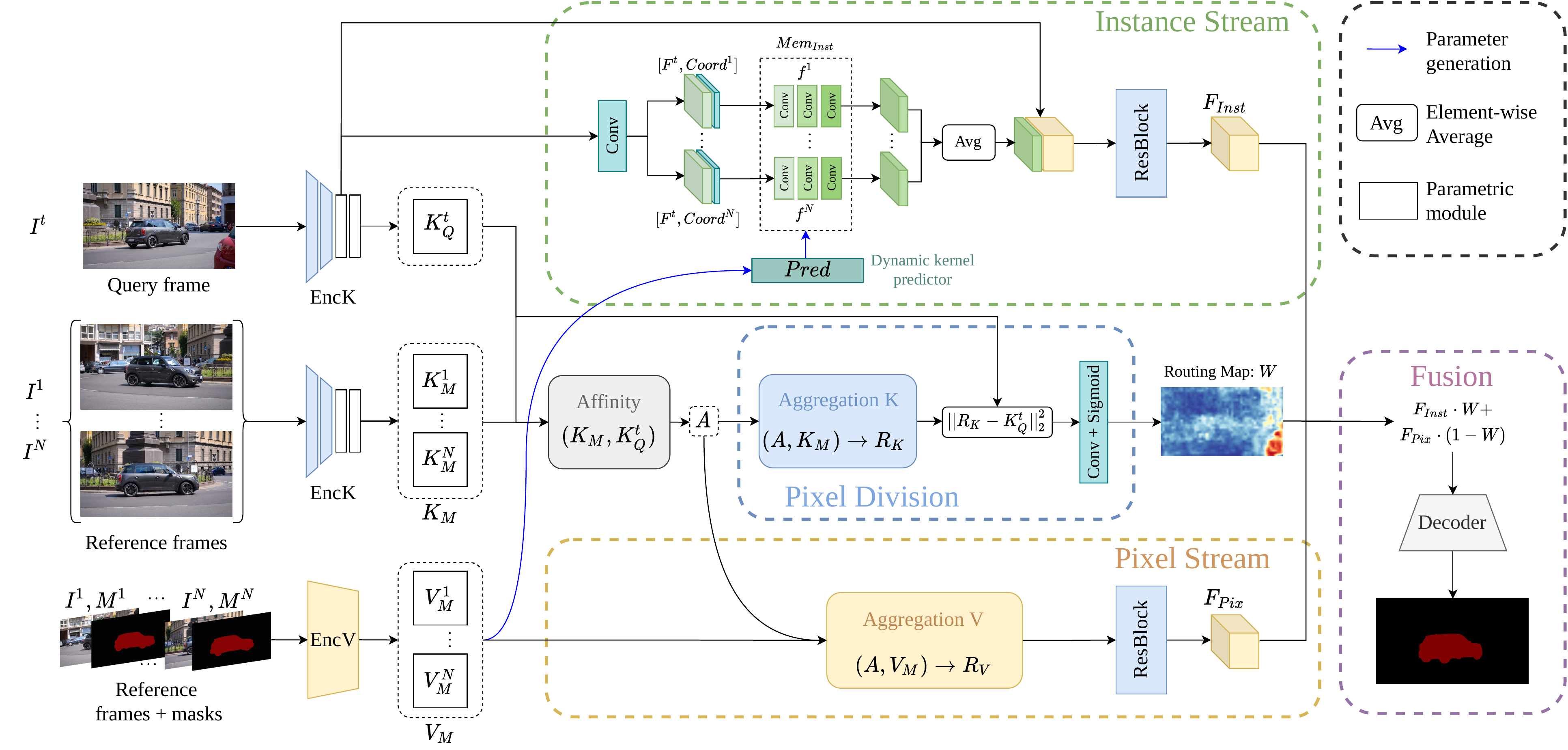} 
  \caption{
    The overview of Two-Stream Network (TSN), which consists of an instance stream, a pixel stream, a pixel division module and a fusion module.
    The instance stream and the pixel stream process the query frame respectively. 
    The pixel division module generates a routing map to seperate the seen and the unseen pixels, and the fusion module merges the output embeddings of the two streams weighed on the routing map.
    }
  \label{fig_overview}
\end{figure*}

\subsection{Overview of the Two-Stream Network}

Given a video sequence of length $T$ and the mask annotation of its first frame, semi-supervised VOS aims to segment the subsequent frames sequentially.
For the current frame $I^t$, \ie, query frame, a reference set $\mathcal{S} = \{ (I^{i}, M^{i}) \}_{i=1}^{N}$ is usually constructed with $N$ historical frames and their corresponding masks.
Under the semi-supervised VOS setting, we propose a Two-Stream Network (TSN).
As illustrated in Figure~\ref{fig_overview}, the proposed TSN includes an instance stream, a pixel stream, a pixel division module and a fusion module.
The pixel stream is deployed to segment the query frame with a conventional pixel-level memory, in which the reference features are retrieved through affinity calculation and feature aggregation with features of query frame.
The instance stream is specially designed to tackle with those \textit{unseen} pixels which lack correspondance in the memory. 
Toward this end, we construct the segmentation heads composed of dynamically generated kernels.
In order to determine the \textit{unseen} pixels, we present the pixel division module. 
It generates a routing map based on the difference between the aggregated feature and the feature of the query frame, which is then leveraged to select pixels without correspondance in the memory.
Finally, the fusion module merges the output embeddings of the two streams with the routing map, and decodes to a segmentation result.
In the following sections, we describe each module in details.

\subsection{Pixel Stream}

The pixel stream is responsible for segmenting the \textit{seen} pixels, for they are able to retrieve reliable correspondences from a pixel-level memory to support segmentation.
Following STCN~\cite{STCN}, a conventional pixel-level memory $\{K_M, V_M\}$ is constructed with reference frames and their masks in the pixel stream.
For memory retrieval, a pixel-wise affinity is calculated and then used to aggregate reference features.
Finally, A mask embedding carrying label information is generated.

Specifically, key encoder $EncK$ is applied to encode the query frame $I^t$ and the reference frames into key features $K_{Q}=\{K_{Q}(p)\}\in \mathbb{R}^{HW\times C_k}$ and $K_{M}=\{K_{M}(q)\}\in \mathbb{R}^{NHW\times C_k}$,
where $p$ and $q$ are the spatial locations, $N$ is the number of frames in the reference set.
Spatial dimensions $H=sH_{im}$ and $W=sW_{im}$, where $H_{im}$ and $W_{im}$ corresponds to the spatial dimension of the image and $s$ to the stride of the backbone in TSN.
The affinity matrix $A\in \mathbb{R}^{HW\times NHW}$ is then calculated through non-local matching between the query key and the reference key.
For each $p$ and $q$, the affinity is calculated as,
\begin{equation}
\label{sim_calc}
  A(p, q)=\frac{\mathrm{exp}(\mathcal{C}(K_Q(p),K_M(q)))}{\sum_{q}{\mathrm{exp}(\mathcal{C}(K_Q(p),K_M(q)))}}
\end{equation}
$\mathcal{C}$ is a similarity function, where we adopt negative squared Euclidean distance as STCN~\cite{STCN}.
Support feature $R_{V}\in \mathbb{R}^{HW\times C_v}$ is aggregated through weighted sum of the value feature $V_M$ of reference frames with the matching affinity $A$,
\begin{equation}
  \label{eq_mem_read}
  R_{V}(p) = \sum_{q}{(A(p, q)\cdot V^M(q))}
\end{equation}
The aggregated feature $R_{V}$ is then concatenated with $F^t$ and processed with a residual block, obtaining the mask embedding $F_{Pix}\in \mathbb{R}^{HW\times C_v}$ of the pixel stream.

\subsection{Instance Stream}

To address the issue of the \textit{unseen} pixels lacking reliable correspondence in the pixel-level memory, we design the instance stream.
Segmentation heads $\{f^1, ..., f^N\}$ composed of dynamic convolutions are generated by a kernel predictor $Pred$, conditioned on the value feature $\{V^1_M, ..., V^N_M\}$ of each instance of the target object.
The deep feature of the query frame is processed by all heads, and the results are averaged element-wisely to generate the final mask embedding. 
%
%

\begin{figure}[t]
  \centering
  \includegraphics[width=0.45\textwidth]{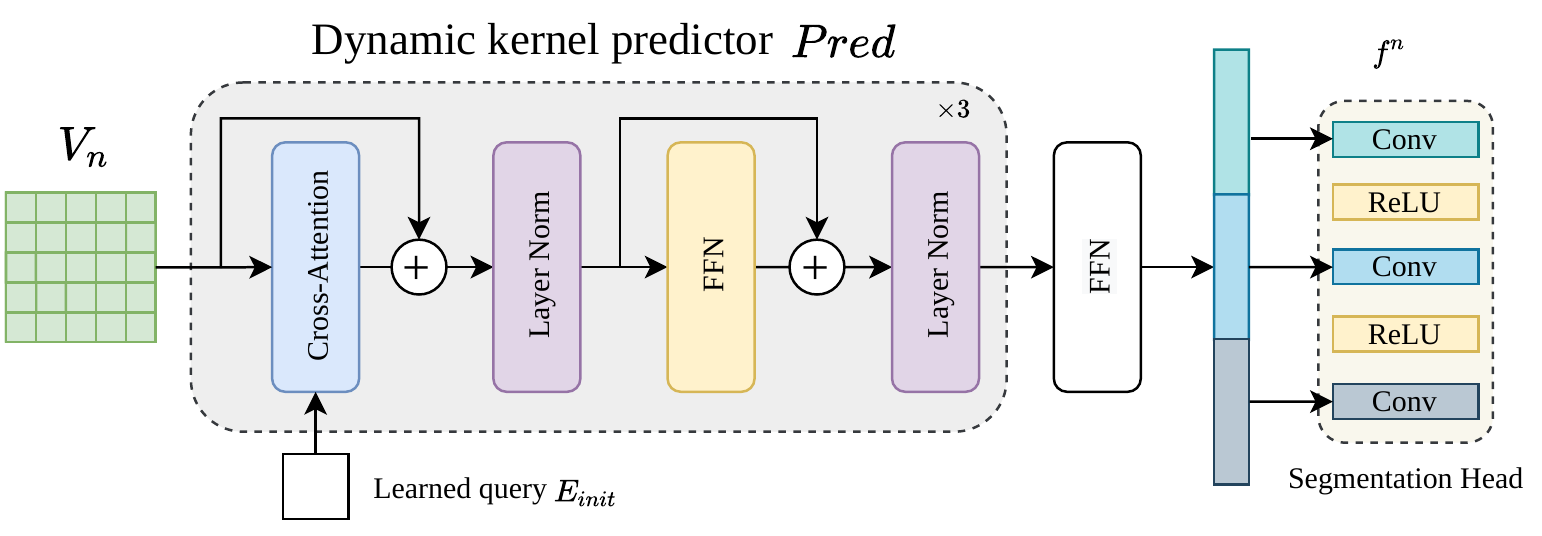} 
  \caption{Details of the dynamic kernel predictor $Pred$, which generates parameters for each segmentation head dynamically.}
  \label{fig_obj_mem_seg}
\end{figure}

Particularly, an instance-level memory $Mem_{inst}=\{f^i\}, i=1, 2, ..., N$ composed of $N$ segmentation heads is constructed.
Each segmentation head is composed of three $1\times 1$ convolution layers with channels of 8, and each layer is activated by a \ReLU~function except for the last one.
The parameters of each segmentation head are dynamically generated from a instance of the target object, characterizing the instance-level cues of the target object, such as the overall appearance and the spatial size.
In practice, we do not generate segmentation heads for the reference frames where objects does not appear.
A dynamic kernel predictor $Pred$ is deployed to generate parameters, taking as input a learnable embedding $E_{init}\in \mathbb{R}^{1\times C_v}$ as well as the value feature of the $n$-th reference frame $V_{M}^n\in \mathbb{R}^{HW\times C_v}$, and outputs a vector $\theta \in \mathbb{R}^{1\times K}$ for each segmentation head, 
The value feature $V_{M}^n\in \mathbb{R}^{HW\times C_v}$ is encoded by value encoder $EncV$, taking as input the reference frame $I^n$ and its corresponding mask $M_n$. 
As shown in Figure~\ref{fig_obj_mem_seg}, $Pred$ is composed of three transformer layers as \cite{transformer}, where $E_{init}$ learns to gather the object information from $V_{M}^n$ adaptively.
The output vector $\theta \in \mathbb{R}^{1\times K}$ can be seen as the parameters of all three convolution layers in a segmentation head concatenated together.

The instance stream segments the query frame $I^t$ with the constructed instance-level memory, taking as input its deep feature $F^t\in \mathbb{R}^{HW\times C_v}$, and outputs mask embedding $F_{Inst}\in \mathbb{R}^{HW\times C_v}$.
The $N$ segmentation heads in $Mem_{Inst}$ segment $F^{t}$ respectively and average their results together as follows,
\begin{equation}
  \label{form_obj_seg}
  O_{Inst} = \frac{1}{N}\cdot \sum_{i}{f^{i}\left ([\mathrm{w}(F^t), Coord_i]\right )}
\end{equation}
where $\mathrm{w}$ is a linear layer used to reducing the channel of $F^t$ from 512 to 8, and $[,]$ represents concatenation along the channel dimension.
$Coord_i$ is a relative coordinate map taking the centroid of the object in reference frame $I^n$ as origin.
Through concatenating the $Coord_i$, the segmentation heads are capable of memorizing the position and size information of an instance, increasing their discriminative ability.
The concatenation of $O_{Inst}$ and $F^t$ then goes through a residual block, obtaining the mask embedding $F_{Inst}\in \mathbb{R}^{HW\times C_v}$.

The previous solutions \cite{CFBI,RPCM} generate a single head averaging all the reference frames.
%
We find it may degrade the discrimination ability of the segmentation head, since that it is extremely difficult to memorize varying appearances and spatial information of an object in one head.
%
%
Additionally, the computational costs are practically the same with TSN, since the generation of a single head in \cite{RPCM} requires to attend to a reference feature of size $NHW\times C$ every time.
%
%
%
%
We provide the quantitative comparison of these two solutions in Table~\ref{tb_ablation_obj_mem}.

\subsection{Pixel Division and Fusion}

The pixel division module is deployed to classify the query pixels into \textit{seen} and \textit{unseen}.
A routing map is generated by the pixel division module based on the difference between the aggregates key features and the feature of the query pixels.

%
%

Specifically, feature $R_K=\{R_K(p)\}\in \mathbb{R}^{HW\times C_k}$ is aggregated through weighted sum of the key feature $K_M$ with the affinity matrix $A$ as follows,
\begin{equation}
  \label{eq_affinity}
  R_K(p) = \sum_{q}{(A(p, q)\cdot V^K(q))}
\end{equation}
Then, the routing map $W=\{W(p)\}\in \mathbb{R}^{HW\times 1}$ can be generated as,
\begin{equation}
  \label{eq_res_map}
  W = \mathcal{S}(\mathcal{F}(D))
\end{equation}
where $\mathcal{F}$ is a linear layer, $\mathcal{S}$ is the sigmoid function, and $D$ is the $L$2 function calculated as $||R_K - K_{Q}||^2$.
Through the value of $W$, the \textit{unseen} pixels can be distinguished from the \textit{seen} pixels, since they tend to obtain larger residual for lacking of similar correspondence in reference.

The fusion module merges the output embeddings of the two streams weighted on the routing map, and decodes to a final segmentation mask.
Fusion with the routing map can ensure that the segmentation results of the \textit{seen} and the \textit{unseen} pixels come from the pixel stream and the instance stream accordingly.
The fusion process follows,

\begin{equation}
  \label{eq_fuse}
  F = W\cdot F_{Inst} + (1 - W)\cdot F_{Pix}  
\end{equation}
At the end, a decoder of FPN structure is adopted to decode $F$ to the final mask $M^t$.

\section{Experiments}
To verify the effectiveness of the proposed Two-Stream Network (TSN), we train a TSN-R50 which uses ResNet-50~\cite{ResNet} as backbone following STCN~\cite{STCN}, and a TSN-SwinB using the strong swin transformer~\cite{swin} as backbone.
We evaluate our two models on two VOS benchmarks, DAVIS~\cite{DAVIS-2017, DAVIS-2016} and YouTube-VOS~\cite{xu2018youtube}.
Ablation studies are performed on the challenging YouTube-VOS~\cite{xu2018youtube} for in-depth analysis. 
%
In the following sections, we first describe the train and evaluation settings and then compare our TSNs with the state-of-the-art semi-supervised VOS methods.
Finally, the ablation studies are presented.

\subsection{Datasets and Evaluation Metrics}
DAVIS has two versions, DAVIS 2016~\cite{DAVIS-2016} contains 20 videos for validation, where each video has one annotated object instance.
DAVIS 2017~\cite{DAVIS-2017} is a multi-object extension of DAVIS 2016, which provides 60 videos for training and 30 videos for validation.
YouTube-VOS 2018~\cite{xu2018youtube} is a large-scale and challenging dataset for video object segmentation, where the training set and validation set contains 3471 and 474 videos respectively. 
YouTube-VOS 2019 further adds additional videos to validation split.

For DAVIS 2016 and 2017, we report the mean of region similarity $\mathcal{J}$, contour accuracy $\mathcal{F}$ and their average $\mathcal{J}\&\mathcal{F}$ for comparison, which are calculated with the standard DAVIS-2017 evaluation toolkit.
For YouTube-VOS 2018 and 2019, we report $\mathcal{J}$ and $\mathcal{F}$ for both \textit{seen} and \textit{unseen} categories, and the averaged overall score $\mathcal{G}$, which are all obtained from the Codalab server.

\begin{table}[t]
  \centering
  \begin{tabular}{lccccc}
  \toprule
  \multicolumn{6}{c}{\textit{Validation 2018 Split}}    \\
  \midrule
  Method    & $\mathcal{G}$ & $\mathcal{J}_{s}$ & $\mathcal{J}_{u}$ & $\mathcal{F}_{s}$ & $\mathcal{F}_{u}$ \\
  \midrule
  PReMVOS   & 66.9 & 71.4 & 56.5 & 75.9 & 63.7 \\
  A-GAME    & 66.1 & 67.8 & 60.8 & -    & -    \\
  STM       & 79.4 & 79.7 & 72.8 & 84.2 & 80.9 \\
  CFBI      & 81.4 & 81.1 & 75.3 & 85.8 & 83.4 \\
  RMNet     & 81.5 & 82.1 & 75.7 & 85.7 & 82.4 \\
  LCM       & 82.0 & 82.2 & 75.7 & 86.7 & 83.4 \\
  SST       & 81.7 & 81.2 & 76.0 & -    & -    \\
  HMMN      & 82.6 & 82.1 & 76.8 & 87.0 & 84.6 \\
  JOINT     & 83.1 & 81.5 & 78.7 & 85.9 & 86.5 \\
  STCN      & 83.0 & 81.9 & 77.9 & 86.5 & 85.7 \\
  AOT       & 83.8 & 82.9 & 77.7 & \underline{87.9} & 86.5 \\
  RPCM      & 84.0 & 83.1 & 78.5 & 87.7 & 86.7 \\
  TSN-R50(Ours) & \underline{84.8} & \underline{83.6} & \underline{79.8} & \underline{87.9} & \underline{87.8} \\
  TSN-SwinB(Ours) & \textbf{86.1} & \textbf{85.1} & \textbf{80.6} & \textbf{89.7} & \textbf{89.1} \\
  \toprule
  \multicolumn{6}{c}{\textit{Validation 2019 Split}}    \\
  \midrule
  Method    & $\mathcal{G}$ & $\mathcal{J}_{s}$ & $\mathcal{J}_{u}$ & $\mathcal{F}_{s}$ & $\mathcal{F}_{u}$ \\
  \midrule
  CFBI      & 81.0 & 80.6 & 75.2 & 85.1 & 83.0 \\
  SST       & 81.8 & 80.9 & 76.6 & -    & -    \\
  HMMN      & 82.5 & 81.7 & 77.3 & 86.1 & 85.0 \\
  JOINT     & 82.8 & 80.8 & 79.0 & 84.8 & 86.6 \\
  STCN      & 84.2 & 82.6 & 79.4 & 87.0 & 87.7 \\
  AOT       & 83.7 & 82.8 & 78.0 & \underline{87.5} & 86.7 \\
  RPCM      & 83.9 & 82.6 & 79.1 & 86.9 & 87.1 \\
  TSN-R50(Ours) & \underline{84.6} & \underline{83.1} & \underline{80.2} & 87.2 & \underline{87.8} \\
  TSN-SwinB(Ours) & \textbf{86.0} & \textbf{84.8} & \textbf{80.9} & \textbf{89.1} & \textbf{89.2} \\
  \bottomrule
  \end{tabular}
  \caption{
    Qualitative comparison with different methods on YouTube-VOS. 
    Subscripts $S$ and $U$ represents the seen and the unseen category respectively.
    }
  \label{tb_yt}
\end{table}

\begin{table}[t]
  \renewcommand\arraystretch{1.2}
  \setlength\tabcolsep{3.5pt} 
  \begin{tabular}{lccc|ccc}
  \toprule
  \multicolumn{1}{c}{\multirow{2}{*}{Method}} & \multicolumn{3}{c}{\textit{2017}} & \multicolumn{3}{c}{\textit{2016}} \\ \cline{2-7} 
  \multicolumn{1}{c}{}     & $\mathcal{J}\&\mathcal{F}$ & $\mathcal{J}$ & $\mathcal{F}$ & $\mathcal{J}\&\mathcal{F}$ & $\mathcal{J}$ & $\mathcal{F}$ \\ 
  \midrule
  OnAVOS                   & 67.9    & 64.5    & 71.3    & 85.5    & 86.1    & 84.9    \\
  OSVOS                    & 59.2    & 56.6    & 61.8    & 80.2    & 79.8    & 80.6    \\
  RGMP                     & 63.2    & 64.8    & 68.6    & 81.7    & 81.8    & 81.5    \\
  FAVOS                    & 58.2    & 54.6    & 61.8    & 81.7    & 81.0    & 82.4    \\
  CINN                     & 70.7    & 67.2    & 74.2    & 84.2    & 83.4    & 85.0    \\
  VideoMatch               & 62.4    & 56.5    & 68.2    & 81.9    & 81.0    & 80.8    \\
  PReMVOS                  & 77.8    & 73.9    & 81.7    & 86.8    & 84.9    & 88.6    \\
  A-GAME                   & 70.0    & 67.2    & 72.7    & 82.1    & 82.2    & 82.0    \\
  FEELVOS                  & 71.6    & 69.1    & 74.0    & 82.2    & 81.7    & 88.1    \\
  STM                      & 81.8    & 79.2    & 84.3    & 89.3    & 88.7    & 89.9    \\
  KMN                      & 82.8    & 80.0    & 85.6    & 90.5    & 89.5    & 91.5    \\
  CFBI                     & 81.9    & 79.1    & 84.6    & 89.4    & 88.3    & 90.5    \\
  RMNet                    & 83.5    & 81.0    & 86.0    & 88.8    & 88.9    & 88.7    \\
  SST                      & 82.5    & 79.9    & 85.1    & -       & -       & -       \\
  HMMN                     & 84.7    & 81.9    & 87.5    & 90.8    & 89.6    & 92      \\
  STCN                     & 85.3    & 82.0    & 88.6    & \underline{91.7} & \underline{90.4} & 93.0 \\
  AOT                      & 83.8    & 81.1    & 86.4    & 90.4    & 89.6    & 91.1    \\
  RPCM                     & 83.7    & 81.3    & 86.0    & 90.6    & 87.1    & \textbf{94.0}  \\
  TSN-R50(Ours)            & \underline{86.2}    & \underline{82.8}    & \underline{89.6}    & 91.0    & 90.1    & 91.8   \\
  TSN-SwinB(Ours)          & \textbf{87.5} & \textbf{84.0} & \textbf{91.0}   & \textbf{92.2}    & \textbf{90.8}    & \underline{93.5}   \\
  \bottomrule
  \end{tabular}
  \caption{
    Qualitative comparison with different methods on DAVIS. 
    %
    %
    Methods with $FS$ represents tested with full resolution videos instead of 480p.
    \textbf{Bold} and \underline{underline} indicate the best and the second-best performance respectively.
    R50 and SwinB represent adopting ResNet-50~\cite{ResNet} and SwinB\cite{swin} as backbone.
    }
  \label{tb_davis17}
\end{table}

\subsection{Training and Inference}
Follow STM~\cite{STM}, we first pre-train TSN on the synthetic dataset, then conduct main training on video dataset.
Image datasets~\cite{coco} are used to generate synthetic video clips for pre-train, where random affine transformations are adopted for sequence synthesis.
Cut\&Paste strategy is also adopted for data augmentation.
Pre-train stage takes a total of $2\times10^5$ iterations with a constant learning rate of $1\times10^{-5}$.
The training split of DAVIS~\cite{DAVIS-2017} and YouTube-VOS~\cite{xu2018youtube} are used for main training.
For each training sample, three frames are randomly collected from a video sequence, augmented by random affine transformation with difference parameters.
The temporal interval range of frame collecting increases from 1 to 25 in the first $1\times10^4$ iterations, and decreases from 25 to 5 in the last $5\times10^4$ epochs.
The main training takes a total of $2\times10^5$ iterations, where poly learning rate decay with initial value of $4\times10^{-5}$ is adopted.
We use a batch size of 4 and a patch size of 432 in both per-train and main training stage.
It takes about 32 hours to finish the two stage training with 4 Tesla A100 GPUs.
The parameters of BatchNorm layers in both key and value encoders are frozen for the whole training.
We adopt Adam~\cite{adam} with standard momentum for the optimization of TSN-R50, and AdamW~\cite{adamw} with weight decay of $1\times10^{-4}$ for TSN-SwinB.
To stablize training, Exponential Moving Average (EMA)~\cite{EMA} is used for TSN-SwinB.
The overall loss function of TSN is a combination of bootstrapped cross-entropy loss and mask IoU loss~\cite{mei2021transvos}.
Bootstrapped cross-entropy loss is calculated following STCN~\cite{STCN}, and mask IoU loss is defined as,
\begin{equation}
    \mathcal{L}_{mIoU}(P_i, G_i) = 1 - \frac{\sum_{p\in \Omega}\min (P^p_i, G^p_i)}{\sum_{p\in \Omega}\max (P^p_i, G^p_i)}
\end{equation}
where $P$ and $G$ is the predicted mask and ground-truth mask of object $i$, $\Omega$ represents all pixels in mask $P$ and $G$.
%
%
Top-$k$ strategy with $k=20$ is adopted during inference as STCN~\cite{STCN}.

\begin{figure*}[t]
  \centering
  \includegraphics[width=1.0\textwidth]{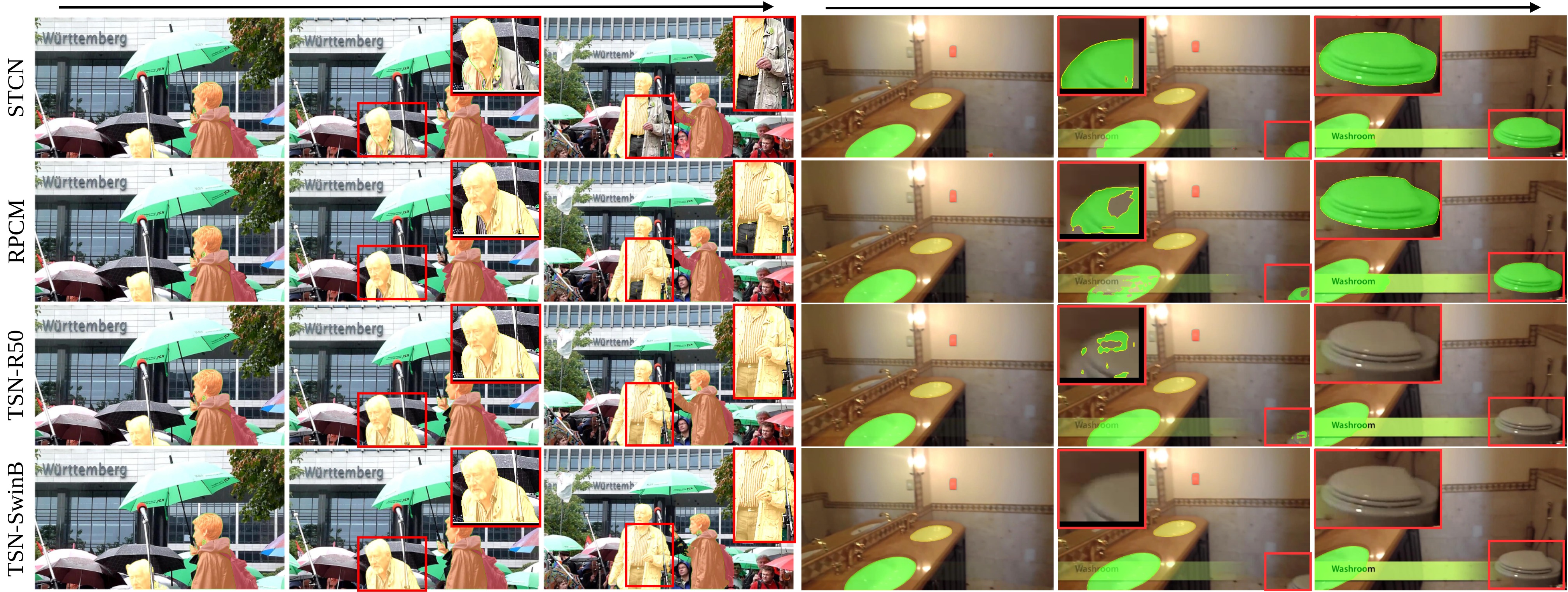} 
  \caption{
    Visual comparison of segmentation results on \textit{6ae0cac484} and \textit{62bf7630b3} of the challenging YouTube-VOS. 
    We present the results of STCN~\cite{STCN} and RPCM~\cite{RPCM} for comparison.}
  \label{fig_sota_res}
\end{figure*}

\subsection{Comparison with State-of-the-arts}
%
The quantitative results of DAVIS 2016 and 2017~\cite{DAVIS-2016, DAVIS-2017} are shown in Table~\ref{tb_davis17}.
%
%
On DAVIS 2017 validation set, which is the multi-object extension of DAVIS 2016, the TSN achieves a $J\&F$ score of 87.5\% when testing with 480p resolution videos, which is new state-of-the-art.

The quantitative results of YouTube-VOS 2018 and 2019 validation sets are presented in Table \ref{tb_yt}.
Due to the large amount of test videos, YouTube-VOS poses a huge challenge to VOS approaches.
Our TSR achieves state-of-the-art $\mathcal{J}\&\mathcal{F}$ performance on both 2018 and 2019 split, where the averaging scores $\mathcal{G}$ are 86.1\% and 86.0\%, respectively.
Visual comparisons are provided in Figure~\ref{fig_sota_res}.
As demonstrated in video \textit{6ae0cac484}, STCN~\cite{STCN} and RPCM~\cite{RPCM} have errors segmenting the emerging person in the box. 
%
In contrast, our TSR segments the emerging area more precisely owing to the two-stream network.
The segmentation ability of our TSR is also demonstrated in video \textit{62bf7630b3} for successfully avoid the disturbance brought by the emerging object in the background.
More visual results will be provided in the supplementary material for comparison.

\subsection{Ablation study}

To evaluate the contribution of each component in the proposed TSN, we conduct ablation study on the Youtube-2018 validation set.
%
In the following, all comparison experiments are conducted based on the TSN-R50 for time efficiency, and models are trained with the video datasets only.

\begin{table}[t]
  \centering
  \setlength\tabcolsep{4pt} 
  \begin{tabular}{ccc|ccc}
  \toprule
  $N_{Head}$ & $N_{Depth}$ & Predictor & $\mathcal{J}\&\mathcal{F}$ & $\mathcal{J}_{seen}$ & $\mathcal{J}_{unseen}$ \\
  \midrule
  N     & 3      & $Pred$      & \textbf{83.8} & \textbf{82.8}  & \textbf{78.4}    \\
  1     & 3      & $Pred$      & 83.0 & 82.2  & 77.0    \\
  N     & 3      & $\mathrm{GAP}$       & 83.5 & 82.3  & 78.2    \\
  N     & 1      & $Pred$      & 83.2 & 82.3  & 77.3    \\
  \bottomrule
  \end{tabular}
  \caption{Ablation on several design choices of the instance stream. 
  $N_{Head}$ represents the number of heads in the instance-level memory.
  $N_{Depth}$ represents the depth of each head.
  $\mathrm{GAP}$ represents global average pooling.}
  \label{tb_ablation_obj_mem}  
\end{table}

\noindent{\textbf{Instance stream}.}
We further investigate several design choices for the instance stream.
Existing solutions~\cite{CFBI, RPCM} utilize object-level information for VOS.
In \cite{CFBI, RPCM}, an object embedding is generated through global pooling on all the reference frames together, which is then used to re-weight the query feature.
For comparison, we conduct ablation experiments on the constructed instance-level memory.
Model$1$ utilizes the proposed instance-level memory, whose segmentation heads are generated from each instance of the target object separately with the proposed predictor.
Model$2$ aggregates feature from all reference frames together generating one segmentation head.
In model$3$, parameter predictor $Pred$ is replaced with a global average pooling layer.
In model$4$, the output of $Pred$ is multiplied directly onto the query feature, rather than forming a segmentation head with multi-layer convolutions.
As shown in Table~\ref{tb_ablation_obj_mem}, the $\mathcal{J}\&\mathcal{F}$ of model$2$ drops from 83.7 to 83.0.
We speculate that mixing all reference together may lead to information confusion, which damages the discrimination capacity of the segmentation heads.
The $\mathcal{J}\&\mathcal{F}$ also drops when replacing $Pred$ with global average pooling, indicating the predictor gathering information with attention learns a better object representation.
Model$1$ achieves better performance than model$4$, proving the superiority of the dynamically generated convolutions.

\begin{table}[t]
  \centering
  \begin{tabular}{c|ccc}
    \toprule
    Position map & $\mathcal{J}\&\mathcal{F}$ & $\mathcal{J}_{seen}$ & $\mathcal{J}_{unseen}$ \\
    \midrule
    -            & 83.0 & 82.1  & 77.5    \\
    Sine         & 83.5 & 82.3  & 77.6    \\
    Rel coord    & \textbf{83.8} & \textbf{82.8}  & \textbf{78.4}    \\
    \bottomrule
  \end{tabular}
  \caption[]{
    Ablation on the design choices for position map.
    Sine: the sine position encoding in \cite{transformer}.
    Rel coord: relative coordinate map in \ref{form_obj_seg}.
    }
  \label{tb_ablation_pos}
\end{table}

\noindent{\textbf{Position map}}
When segmenting with the instance-level memory, we take as input the concatenation of the query feature and a relative position map, which provides a strong cue for segmentation.
We also conduct ablation study to investigate the design choices of the position map.
As shown in Table~\ref{tb_ablation_pos}, removing position map leads to a significant performance drop.
We conjecture that through the relative position, the segmentation head can better realize the spatial information of the object, such as spatial position and size.
Replacing the relative position map with the absolute sine position encoding also decrease the performance slightly.
We speculate that the normalized relative coordinates are easier to recognize than the sine position encoding, for each head has only 8 channel.

\begin{figure}[t]
  \centering
  \includegraphics[width=0.45\textwidth]{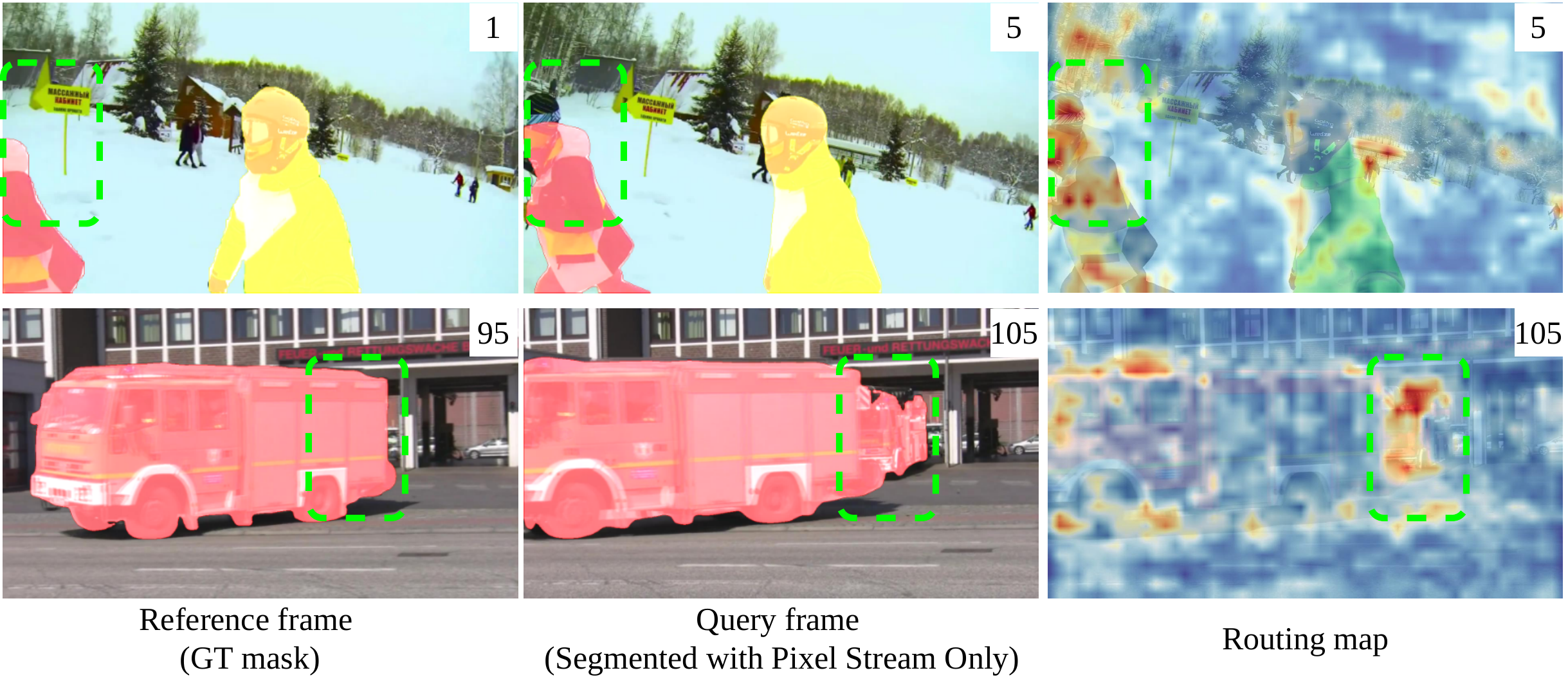} 
  \caption{
    Visualization of the routing map.
    The area in the dashed box on the query frame is the unseen area in the reference frame.
    They have relatively large values on the routing map (heat area).
    }
  \label{figure_rmap}
\end{figure}

\begin{figure}[t]
  \centering
  \includegraphics[width=0.39\textwidth]{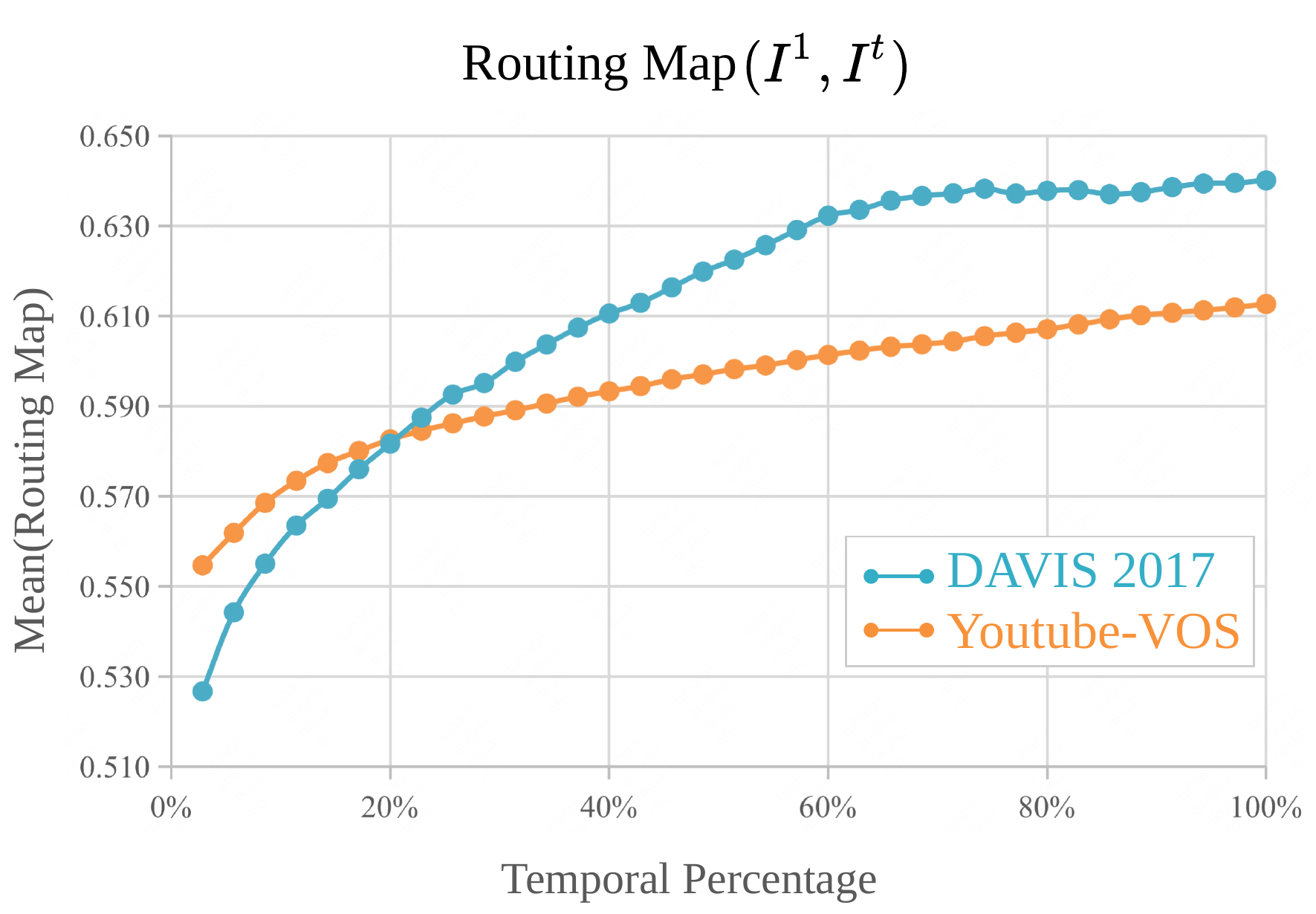} 
  \caption{
    Mean value of routing maps, calculated with the difference between the $t$-th frame $I^t$ and the first annotated frame $I^1$.
    The value of routing map increase over time, indicating that unseen pixels are gradually emerging along the video.
    }
  \label{fig_rmap_static}
\end{figure}

\begin{table}[]
  \centering
  \begin{tabular}{cc|cccc}
    \toprule
    $\mathrm{Inst}$   & $\mathrm{Pix}$   & $\mathcal{J}\&\mathcal{F}$ & $\mathcal{J}_{seen}$ & $\mathcal{J}_{unseen}$ & FPS \\
    \midrule
    $W$   & 1-$W$ & \textbf{83.8} & \textbf{82.8}  & \textbf{78.4}  & 17.2  \\
    0   & 1   & 82.6 & 82.1  & 76.5  & 18.6  \\
    0.5 & 0.5 & 82.8 & 81.5  & 77.2  & 17.2  \\
    1   & 0   & 55.4 & 62.8  & 45.5  & \textbf{25.0}  \\
  \bottomrule
  \end{tabular}
  \caption[]{
    Ablation on the two stream of TSN.
    $\mathrm{Inst}$ and $\mathrm{Pix}$ represent the instance stream and the pixel stream, while their value indicates the way of combining the two streams. 
    1st: combining the two streams with the generated routing map.
    2nd: only the pixel stream.
    3rd: combining the two streams with equal weights, which are spatial-invariant.
    4th: only the instance stream.
    }
  \label{tb_ablation_rmap}
\end{table}

\noindent{\textbf{Pixel division module}.}
Being a critical module in the two-stream network, pixel division module splits the query pixels into \textit{seen} and \textit{unseen} through a routing map.
%
To investigate the dividing ability of the pixel division module, a visualization of the routing map is presented in Figure~\ref{figure_rmap}.
Pixels that have been unseen in the reference frame appear in the query frame, while the pixel stream is unable to classify them correctly.
On the routing map, the \textit{unseen} pixels have relatively large values, which makes their segmentation results mainly determined by the instance branch in our TSN.
%
Figure~\ref{fig_rmap_static} plots the mean values of routing maps calculated by the first annotated frame ($I^1$) and each subsequent frame ($I^t$).
The mean values which characterize the differences between each two frames raise along the video.

\begin{figure}[t]
  \centering
  \includegraphics[width=0.49\textwidth]{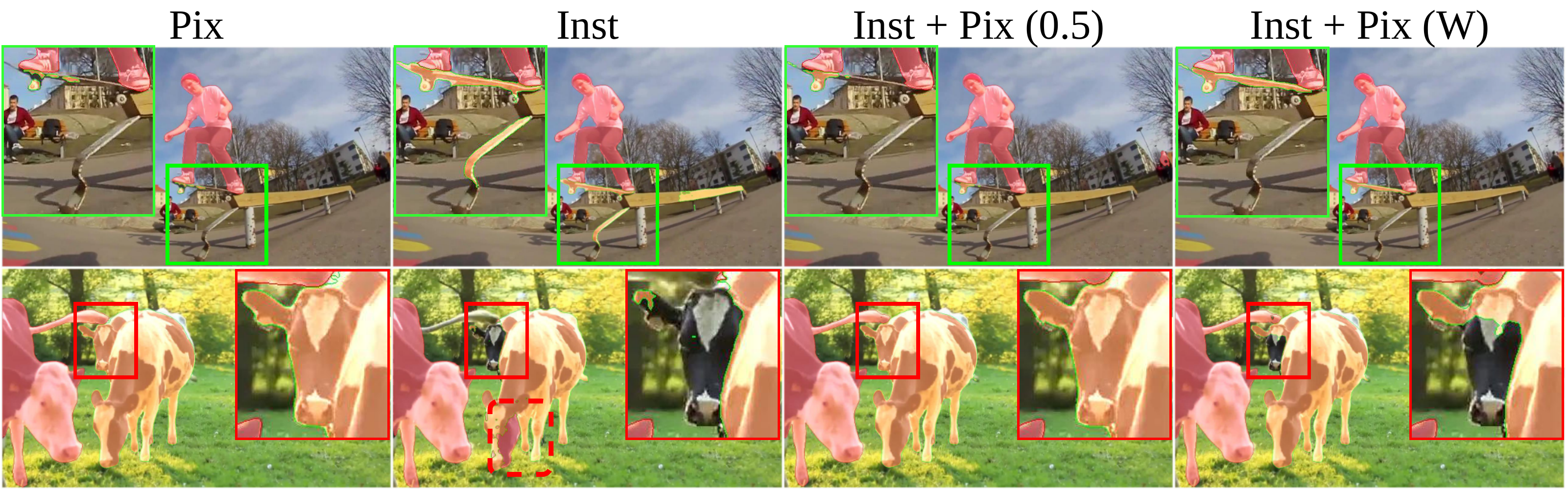} 
  \caption{
    Visual comparison of the proposed two streams.
    Each stream owns its inherent characteristics.
    Simply fusing them with equal weights is not enough to combine their strengths.
    }
  \label{figure_two_stream}
\end{figure}

\noindent{\textbf{Fusion with Routing Map}.}
The output embeddings are fused weighted on the generated routing map.
%
We provide quantitative results to verify the effectiveness of the proposed fusion strategy.
Models with only instance stream, only pixel stream and combining the two streams with an equal weight are trained for comparison.
As shown in Table~\ref{tb_ablation_rmap}, model fusing the two streams with the routing map achieves the best results.
%
Performance of the model with only pixel stream drops since its lack the ability to deal with the \textit{unseen} pixels.
Combining the two streams with an equal weight also leads to performance drop, because they may interfere with each other in areas where they are not skilled.
The performance drops severely when leaving only instance stream, indicating TSR still heavily relies on the pixel-level cue.
Visual comparison of the two streams are provided in Figure~\ref{figure_two_stream}, presenting that the pixel stream and the instance stream owns their inherent characteristics.
%
Simply fusing them with equal weights is not enough to combine their strengths.

\section{Conclusion}

We have presented a Two-Stream Network for video object segmentation task.
%
Our work shows that it is greatly beneficial in VOS to separate the pixels of the query image into unseen and seen pixels with a routing map, and process them with the object-wise and pixel-wise segmentation module respectively.
We also propose a novel dynamic kernel-based module to obtain the instance-level segmentation, which can encode the target object in its weights and efficiently and accurately segment it in the query image. 
Our framework achieves the new state-of-the-art performance on both DAVIS and the large-scale YouTube-VOS and we believe that the proposed framework is a simple and strong baselines for further research.

\bibliographystyle{abbrv}
\bibliography{bibtex}
\end{document}